\documentclass[10pt,twocolumn,letterpaper]{article}

\usepackage{wacv}
\usepackage{times}
\usepackage{epsfig}
\usepackage{graphicx}
\usepackage{amsmath}
\usepackage{amssymb}
\usepackage{booktabs}
\usepackage[font=footnotesize]{subfig}

\def\wacvPaperID{****} 

\wacvapplicationstrack 

\wacvfinalcopy 

\ifwacvfinal
\usepackage[breaklinks=true,bookmarks=false]{hyperref}
\else
\usepackage[pagebackref=true,breaklinks=true,colorlinks,bookmarks=false]{hyperref}
\fi

\begin{document}

\title{Expanding Accurate Person Recognition to New  Altitudes and Ranges: The BRIAR Dataset}
\author{
David Cornett III \and 
Joel Brogan \and
Nell Barber \and

Deniz Aykac \and
Seth Baird \and
Nick Burchfield \and
Carl Dukes \and
Andrew Duncan \and 
Regina Ferrell \and
Jim Goddard \and
Gavin Jager \and
Matt Larson \and
Bart Murphy \and
Christi Johnson \and
Ian Shelley \and
Nisha Srinivas \and
Brandon Stockwell \and
Leanne Thompson \and
Matt Yohe \and 
Robert Zhang \and

Scott Dolvin \and
Hector J. Santos-Villalobos \and
David S. Bolme \\ \and

Oak Ridge National Laboratory\\ 
P.O. Box 2008. Oak Ridge, TN 37831\\
{\tt\small \{cornettdciii,bolmeds\}@ornl.gov}


}

\maketitle
\thispagestyle{empty}

\begin{abstract}
    Face recognition technology has advanced significantly in recent years due largely to the availability of large and increasingly complex training datasets for use in deep learning models. 
   These datasets, however, typically comprise images scraped from news sites or social media platforms and, therefore, have limited utility in more advanced security, forensics, and military applications. These applications require lower resolution, longer ranges, and elevated viewpoints.  
    To meet these critical needs, we collected and curated the first and second subsets of a large multimodal biometric dataset designed for use in the research and development (R\&D) of biometric recognition technologies under extremely challenging conditions. Thus far, the dataset includes more than 350,000 still images and over 1,300 hours of video footage of approximately 1,000 subjects. To collect this data, we used Nikon DSLR cameras, a variety of commercial surveillance cameras, specialized long-rage R\&D cameras, and Group 1 and Group 2 UAV platforms. The goal is to support the development of algorithms capable of accurately recognizing people at ranges up to 1,000 m and from high angles of elevation.
    
    These advances will include improvements to the state of the art in face recognition and will support new research in the area of whole-body recognition using methods based on gait and anthropometry.  
    This paper describes methods used to collect and curate the dataset, and the dataset's characteristics at the current stage.
\end{abstract}


\section{Introduction}
\label{sec:intro}
The Biometric Recognition and Identification at Altitude and Range (BRIAR) program was conceived to advance the current state of the art in the areas of face, gait and whole-body (WB) biometric recognition under challenging, uncontrolled condition in full-motion video. This includes supporting capabilities such as person detection and tracking, image enhancement, and atmospheric turbulence mitigation. To achieve these goals, the BRIAR program is building a one-of-a-kind dataset to foster the creation of algorithms to address these challenging scenarios. The BRIAR program will be implemented in several phases, where each phase includes new data collections to add to this dataset, as well as research phases that aim to produce models trained on this new dataset with ever-increasing performance goals. The individual data collections are specifically planned to span varying terrain, weather, and atmospheric conditions and to bolster the demographic diversity of the subject pool. During the first phase of the program, BRIAR Government Collections 1 (BGC1) and 2 (BGC2) were designed to support the development of algorithms targeting these problems. 

Real-world implementations of existing algorithms for face and WB recognition commonly ingest video media from commercial security cameras. Data captured by these sensors is of varying quality, which can degrade algorithm performance and represents a significant shortcoming in biometric recognition\cite{januspaper}. To provide a truly eclectic dataset in terms of imaging sources, the BRIAR dataset will contain image and video data captured not only by commercial surveillance cameras but also by specialized sensors capable of obtaining high-quality video footage of subjects at ranges of up to 1,000 m and at a view angle of up to 50\textdegree. 

Many promising applications of biometric algorithms are further limited by the absence of training data captured from sensors mounted on unmanned aerial vehicle (UAV) platforms. Such platforms are becoming increasingly present in commercial, industrial, and defense applications. To push the boundaries of newer algorithms, the BRIAR BGC1 and BGC2 datasets will also include face and WB video footage taken from altitudes of up to 400 m from four different platforms, both stationary and changing position along three axes over the course of a single recording.

\subsection{Contributions}

The BRIAR dataset makes makes two significant contributions. First, it establishes a one-of-a-kind dataset comprising still images and videos of subjects from multiple ranges and elevations across two sets of clothing. Second, it enables researchers to expand the current state-of-the-art algorithms by leveraging the uniqueness of the dataset and the unexplored imaging challenges that it provides.

Several key features make this dataset truly novel.  Although many existing datasets provide images and/or videos that are either exclusively constrained or exclusively unconstrained\cite{LFWTechUpdate}\cite{liu2015faceattributes}, the BRIAR dataset contains high-quality close-up images and variable-quality unconstrained videos affected by atmospheric conditions (e.g., turbulence, rain, shadow).  Videos of each subject can range from 4K quality at less than 10 m to high definition at distances up to 1, 000 m and include representations of both cooperative and noncooperative behavior. 

The unique features of this dataset will allow IARPA's R\&D teams to advance the current state of the art in the areas of face and WB detection and recognition and person tracking and reidentification. At this time, BRIAR datasets are available to US Government (USG) researchers and their contractors upon request.  Interested parties should contact the authors, who will forward the request to the appropriate USG representatives.

\subsection{Paper Organization}

The remainder of this paper is organized as follows: In Section  \ref{sec:background}, we provide background information on the limitations that the BRIAR dataset is intended to address and discuss several key considerations in the dataset's development. 
Section \ref{sec:prev_work} provides an examination of related work and summarizes the unique goals challenges related to the collection of this dataset. In Section \ref{sec:data_collection}, we discuss the dataset collection protocols and methods, the nature of the unconstrained imaging used in the collection, and details of the collection sites. In Section \ref{sec:dataset}, we describe data post-processing methods, metadata, and means of annotation. In Section \ref{sec:characteristics}, we provide characteristics of the dataset and a discussion of its limitations in its current state. Finally, Section \ref{sec:conclusion} presents a road map for future expansion of the BRIAR dataset.

\section{Background}
\label{sec:background}




Biometric Identification at range or from extreme pitch angles is an important area of continuing research, and has many use cases to society at large. The USG is approved to apply such capabilities as measures towards supporting counter-terrorism, protection of critical infrastructure and military forces, and border security. That being said, although there is a great need for such technology, the current state of the art does not adequately address these extreme imaging scenarios.  

\subsection{Challenges at Altitude and Range}

Deep learning has pushed the state of the art in face recognition such that real successes can be achieved even when addressing unconstrained images of individuals. As noted in \cite{9432821}, this success in so-called in-the-wild face recognition (e.g., various poses, expressions, angles, and lighting conditions) is due in part to both the advances of convolutional neural network--based deep learning methods as well as the curation of very large training datasets. Although recent deep learning techniques have strengthened unconstrained face recognition, the nature of this particular dataset presents unique challenges that remain unsolved by the research community. 

As the distance to the subject increases, the available number of pixels across the face that can be used to form a facial template decreases. When available, WB imagery inherently contains more pixel data than facial imagery at a given standoff distance. At greater distances where stable facial information becomes unattainable  due to lack of pixels, WB biometrics may still be provide viable signatures from an individual. To address these WB imaging considerations in the BRIAR BGC collections, we have selected an extensive set of cameras with various resolutions and optical characteristics, from low-end sensors to high-end diffraction limited systems. . Likewise, the extreme pitch angles achieved by both elevated ground systems and UAV-mounted systems allow for the collection and evaluation of face and WB videos at more severe angles than typically common in other unconstrained datasets. To compound the resolution and pitch angle issues for all such face and WB videos and images, atmospheric distortions created from weather conditions during the collection process further increase the challenging nature of imagery within this dataset. 

The BRIAR collection procedure ensures that most subjects will have some ideal and high-quality media suitable for current biometric algorithms. The focus of the program, however, is to capture very challenging media, including extreme ranges up to 1,000 m, very low resolutions, severe elevation angles, and a variety of locations and weather conditions. In particular, the collection process is designed to capture the following challenges per each subject:

\begin{itemize}
    \item Distances: 100 m--1,000 m
    \item Face width (Resolution): 10 pix--32 pix
    \item Body height (Resolution): 50 pix--200 pix
    \item Yaw angles: Full 360\textdegree
    \item Multiple climates, locations, weather conditions (all contribute to atmospheric distortions)
\end{itemize}

The BGC1 dataset was collected using visible-spectrum cameras categorized broadly into three classes: (1) indoor and close-range cameras, (2) mid-range cameras, and (3) specialized cameras. The class of indoor and close-range cameras comprises commercial security cameras that leverage wide fields of view. This class of cameras also includes standard digital single-lens reflex (DSLR) cameras used for passport-like still images. These systems are used at either the indoor, controlled collection (detailed in Section~\ref{subsec:controlled_collection}) or at close distances. Mid-range cameras have larger zoom range, are capable of narrower fields of view, and are often high-end commercial off-the-shelf (COTS) systems. These camera systems are leveraged for facial imaging up to 300 m and WB imaging up to 600 m. Finally, the specialized camera class comprises custom integrated imaging systems using high-end optics and sensors specifically selected to achieve optimal imaging at range. These systems can be adjusted to achieve higher resolutions, variable frame rates, or other imaging optimizations. 

The specialized class of cameras includes long-range and military-grade cameras designed specifically for surveillance from significant distances. These systems are either speciality COTS systems or imaging systems actively used in strategic operations. While BGC1 does not include media captured from long-range or military-grade cameras or sensors stationed at ranges exceeding 500 m, the BGC2 collection did incorporate such systems to produce samples reflecting increasingly challenging imaging conditions. Long-range and military-grade cameras were stationed between 600 m and 1 km from the collection area.

\subsection{Societal Impact}



Biometric identification systems provide vital intelligence to advance the mission of national security. They support efforts to mitigate crime, reduce terrorism, combat human trafficking, improve public safety, and save lives. Despite these advantages, these systems are not without flaws, and they are not invulnerable to misuse. Without adequate controls, they can be used to violate privacy, infringe upon the civil liberties of law-abiding citizens, and suppress political opposition. Certain guiding principles must remain central to organizations and individuals developing and utilizing biometric tools, and these include compliance with all applicable laws, transparency and accountability, equity and objectivity, resiliency, scientific rigor, and the ultimate authority of human judgment \cite{odnireport}.

Equality is an important goal of the program across all individuals, demographic groups, and body types. Diversity of the subject pool is a key goal of data collection efforts and will be of high importance for algorithm evaluations. To ensure proper use of the dataset, access requires approval from IARPA and Institutional Review Board review, and  appropriate measures to protect the privacy of the human subjects have been implemented to ensure data use is ethical and serves US interests; follows applicable US laws, regulations, and policies; and ultimately protects civil liberties.

\section{Related Work}
\label{sec:prev_work}

Multiple large-scale data collections have been published that are focused on the identification task. BRIAR is unique among them due to the sensor-to-subject distances for both ground and aerial cameras. The most relevant datasets are shown in Table \ref{tab:dataset_table}, where a significant number of subject identities were captured, cameras were placed at long distances to the subjects, or aerial views with other relevant details such as appearance changes were present. A few datasets do not state their maximum range or subject-to-sensor distance, so only an estimate is provided. Similarly, the UAV altitudes are missing, but the authors typically use consumer-type drones. Unless a  Certificates of Waiver or Authorization (COA) is granted for a given collection, the maximum elevation of UAV datasets is estimated to be the Federal Aviation Administration (FAA) limit of 400 ft \cite{FAAlimit}.

The IJB-MDF (IARPA Janus Benchmark Multi-domain Force) dataset captured 251 subjects in indoor and outdoor settings at various ranges and with various camera types. Mugshot-style photos with frontal and side views and at pitch angles of $\pm{15}$ degrees. The longest visible camera was stationed 500 m from the subjects. The data collection made use of multiple IR cameras, but they were stationed only at close ranges \cite{Kalka_MDF}.

Conversely, the UG2+ dataset and challenge~\cite{vidal2018ug} provides unconstrained video captured from UAVs, gliders, and ground videos, however provides no direct focus on biometric recognition or identification.

Clothing change can cause siginificant degredation in WB recognition, and is an open area of research. The vBOLO dataset~\cite{li2016toward} and the IJB-S (IARPA Janus Benchmark Surveillance Video Benchmark) dataset contains images and video of subjects with considerable appearance change. The IJB-S dataset consists of 202 subjects taken over two weekends. Forty of the participants participated in both weekend collects, so it can be assumed that those subjects had clothing changes.Both IJB-S and vBOLO Videos were taken in a variety of scenes to resemble real-life operational environments including subway station, bus station, embassy, and marketplace. Subjects were also recorded for parts of their routes between the different scenes. Some scenarios featured multiple participants at a time. COTS cameras were used, and an for IJB-S, a UAV was flown over the collection area to record video \cite{Kalka_IJBS}.

 IARPA’s MEVA (Multiview Extended Video with Activities) contains data from surveillance-type cameras set up in areas such as a school building, bus stop, subway station, and parking areas. The dataset also contains UAV data from a pair of DJI drones and labeled activities. The subjects also carried GPS trackers, which is a unique feature of this dataset\cite{Corona_2021_WACV}.


P-DESTRE, PRAI-1581, RPField, DukeMTMC, and UCCS datasets were all collected on college campuses or public spaces, so those contain a significant number of unique identities. However, they are all lacking in terms of long distances and high elevations \cite{pdestre, gunther_unconstrained_2017, ristani_performance_2016, zheng_rpifield_2018, zhang_person_2021}. The UCCS dataset uniquely contains recordings from different seasons, so there are significant environmental and clothing changes. The P-DESTRE dataset features only clothing changes. PRAI-1581 and P-DESTRE solely consist of UAV imagery, but the drones are flown at less than 60 m. PRAI-1581 only publicly provides WB image crops instead of the full video, so it is labeled as a nongroup dataset in Table \ref{tab:dataset_table}.

The Long-Range Face Identification Dataset (LFRID) (currently unpublished) was collected by the US Army C5ISR (Command, Control, Communications, Computers, Cyber, Intelligence, Surveillance, and Reconnaissance) Center for the purpose of studying face recognition at long ranges up to 1,000 m with significant atmospheric and turbulence effects.  It is ideal for research of long-range identification but, compared to BRIAR, contains fewer subjects (just 149) and locations (just one: hot and arid).  In addition, 49 of the subjects are in a sequestered dataset for US government evaluation only.  

\begin{table*}
\centering
\begin{tabular}{llllll}
\toprule
Dataset     & \# Subjects                             & Max. range      & Max. elevation            & Appearance change & Groups  \\
\midrule
IJB-MDF     & 251                                     & 500 m           & Ground                   & No               & Yes     \\
IJB-S       & 202                                     & Estimated $<$500 m & FAA limit       & Yes (40)               & Yes     \\
RPField     & 112 + \textasciitilde{}4,000 distractors & 158 m           & Cameras on poles & No               & Yes     \\
PRAI-1581   & 1,581                                    &                & 60 m                      & No               & No      \\
MEVA        & 100                                     & Estimated $<$500 m & FAA limit                & No               & Yes     \\
UG$^2$+        & N/A                                     & Estimated $<$500 m & FAA limit                & No               & No     \\
UCCS        & 1,732                                    & 150 m           & Ground                   & Yes              & Yes     \\
DukeMTMC    & 2,700                                    & Estimated $<$100 m & Estimated $<$10 m            & No               & Yes     \\
P-DESTRE &	269	& &	6.7m &	Yes	& Yes \\
LFRID &	100 (+ 49 sequestered)	& 1000m &	Ground &	No	& No \\
BGC1 (ours) & 312 (+ 161 distractors)                                     & 500 m           & 400 m                      & Yes              & No \\
BGC2 (ours) & 302 (+ 280 distractors)                                  & 1,000 m           & 400 m                      & Yes              & No \\
\bottomrule
\end{tabular}
\caption{Comparison to other identification-focused datasets that share similarities to BRIAR.}
\label{tab:dataset_table}
\end{table*}

\section{Data Collection Methods} 

\label{sec:data_collection}

The BRIAR dataset is composed of images and videos captured in a controlled indoor setting and videos captured in an semicontrolled field setting. While most subjects completed the full scope of activities, a subset completed only the controlled portion and are included in BRIAR Test Set 1 and BRIAR Test Set 2 as distractors that help simulate a larger gallery. Regardless of whether subjects were assigned to the full collection or to the distractor pool, in most cases they performed all of their designated activities twice, each time in different clothing. Evaluations will focus on cross--clothing set matching to discourage reliance on traditional Re-ID methods as well as focus algorithms on actual biometric identification. Because it was not always logistically possible to provide private changing areas, participants were instructed to bring additional clothing that they could put on over their first outfit. Those who brought a second pair of shoes also changed footwear, which may provide an additional challenge for gait recognition methods. We requested that subjects remove hats and sunglasses if they were comfortable doing so but permitted them for outdoor activities if subjects were concerned about excessive sun exposure.

For all subjects, hardware time was synchronized using Network Time Protocol services or GPS devices.  Study team members recorded timestamps for when subjects started and stopped activities using a desktop application developed and maintained specifically for BRIAR data collection.  The timestamps were then used during data curation activities to cut videos into segments associated with individual subjects and to associate the media with additional metadata such as weather measurements.  Although most equipment produces timestamps with precision of a few hundredths of a second, some commercial surveillance cameras can have 2--3 seconds of lag or drift, making precise time alignment challenging.  For most metadata, this is not a problem, but it does cause challenges when trying to precisely label activities or some covariates related to subject movement such as yaw and pitch angle estimates.


\begin{figure}[htp]

\subfloat[Top-down view of the BGC1 collection site layout. The physical left-to-right distance spans nearly .5 mi. Labeled areas: (i) intake tent (not shown in Fig.~\ref{fig:BGC2_physical_layout}), (ii) controlled collection tent(not shown in Fig. in~\ref{fig:BGC2_physical_layout}), (iii) dome camera mast, (iv) outdoor collection tent, (v) 100 m range cameras, (vi) 200 m range cameras, (vii) 300 m range cameras, (viii) 400 m range cameras, (ix) 500 m range cameras, (x) weather capture, (xi) UAV control center, (xii) UAV landing pad , (xiii) sensor research]{%
  \includegraphics[clip,width=\columnwidth]{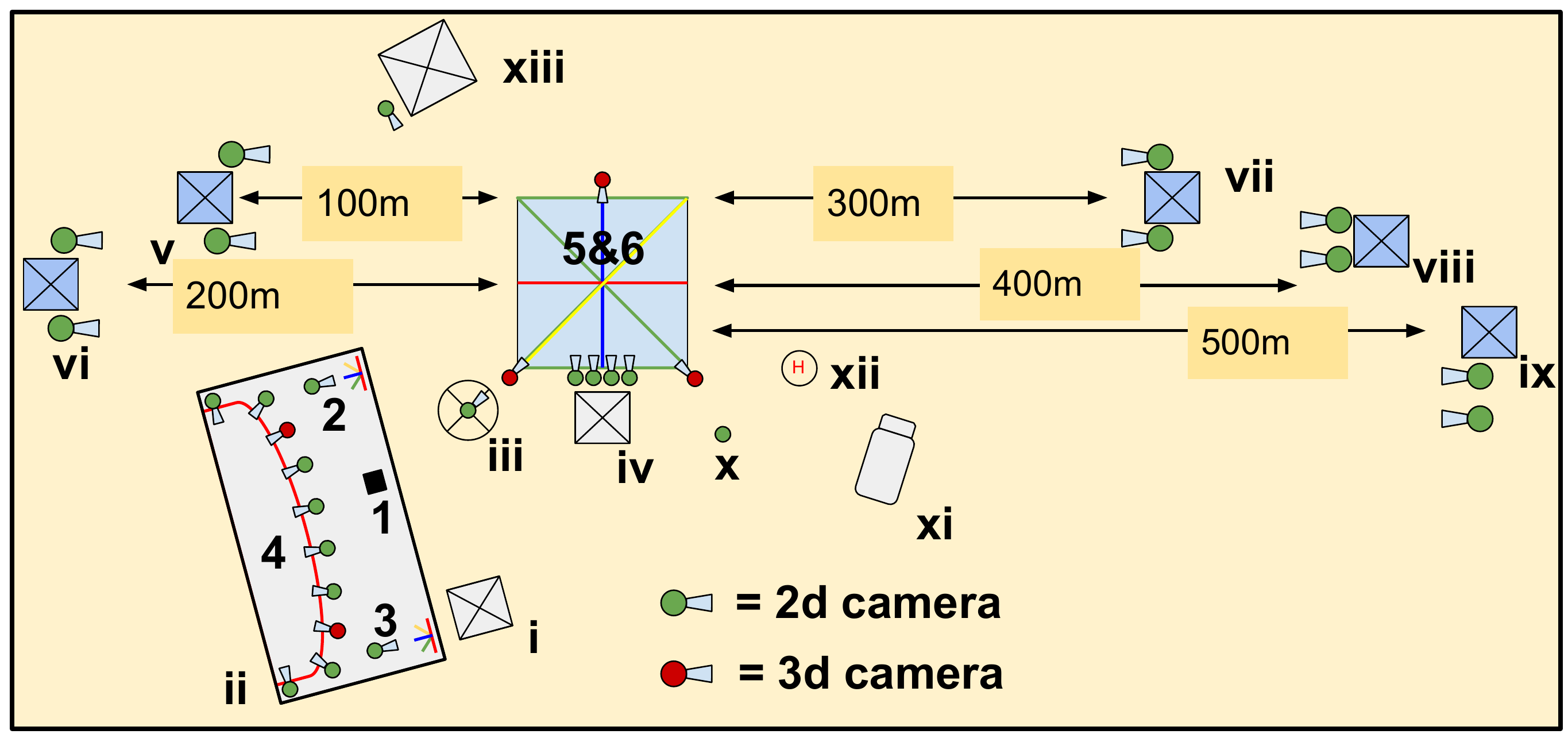}%
  \label{fig:BGC1_physical_layout}
}

\subfloat[Top-down view of the BGC2 collection site layout. Unlike in BGC1 Fig.~\ref{fig:BGC1_physical_layout}, BGC2 intake and control collection areas are located in a separate area Labeled areas: (xiv) 800 m range cameras (added in BGC2), (xv) 1000 m range cameras (added in BGC2)]{%
  \includegraphics[clip,width=\columnwidth]{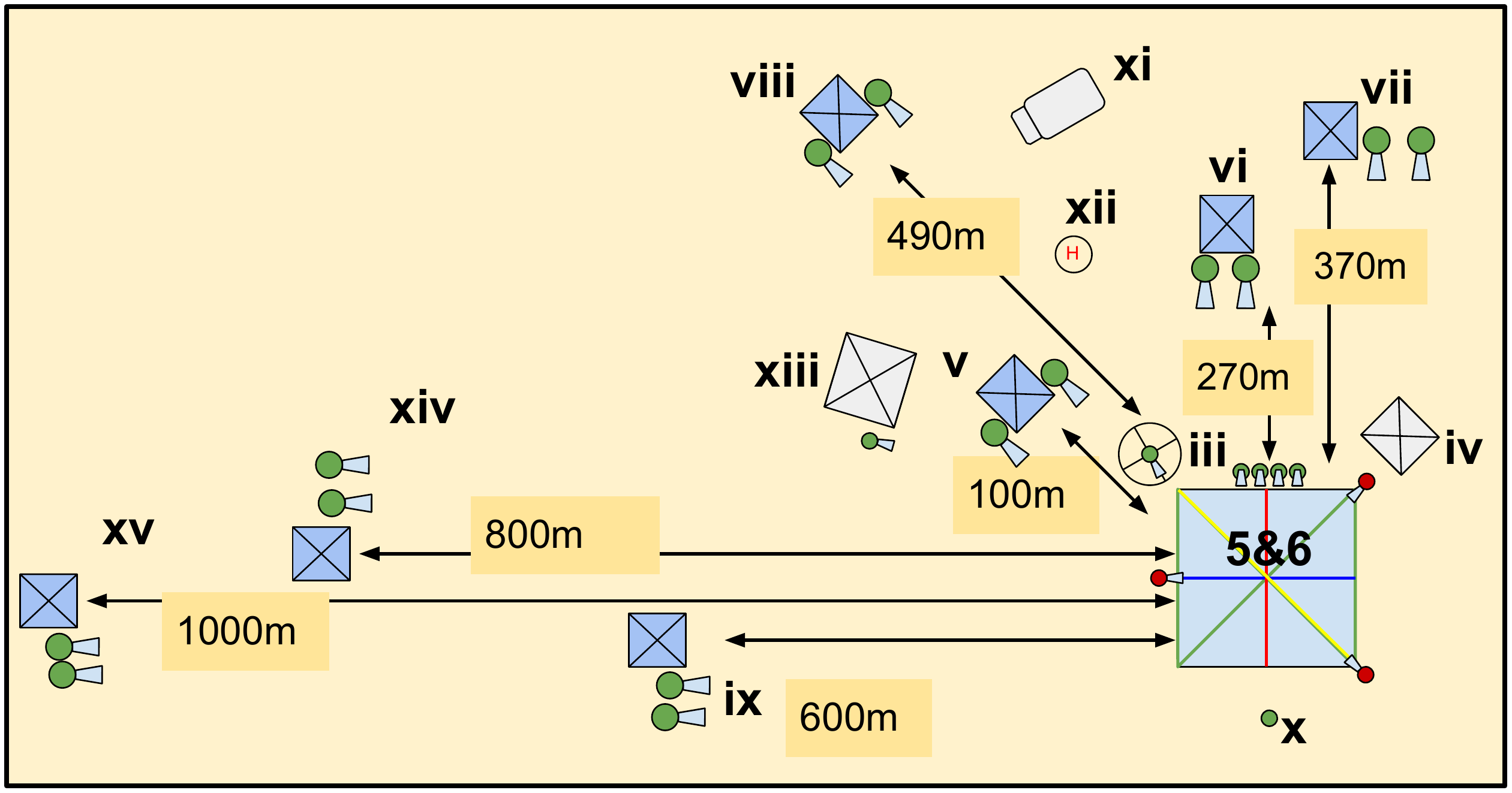}%
  \label{fig:BGC2_physical_layout}
}

\caption{Collection setups for BGC1 and 2. Common Numbered activity stations: (1) height and weight collection, (2) face passport image collection, (3) body passport image collection, (4) controlled walking arena, and (5\&6) outdoor walking arena.}

\end{figure}



\subsection{Controlled Collection}
\label{subsec:controlled_collection}
\begin{figure}
    \centering
    \includegraphics[scale=0.3]{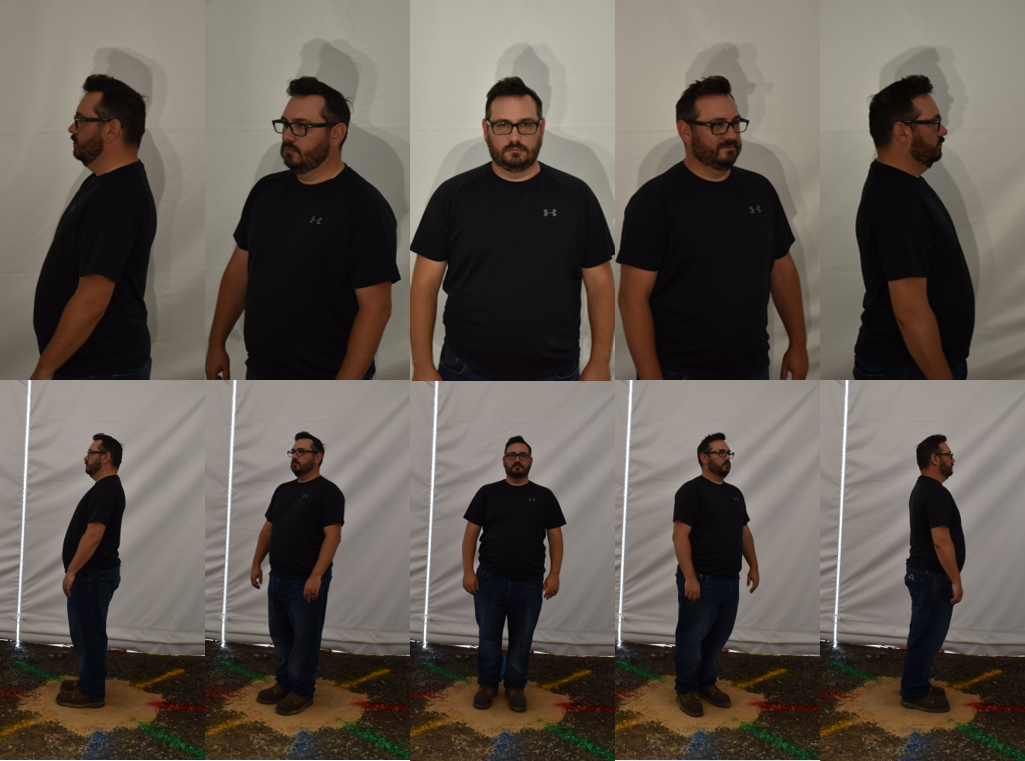}
    \caption{Controlled collection image example of a subject. The top row contains the five neutral face still images. The bottom row displays five of the eight collected angles for standing, WB still images.}
    \label{fig:controlled_example}
\end{figure}

Operationally, biometrics involve the use of constrained images like those used for government-issued IDs or mugshots. To build a set of images and videos in a controlled setting mimicking these conditions, we created three indoor stations: two stations for still images and one station for video-recorded activities. Our procedure is intended to capture more yaw and elevation angles than would typically be captured for a photo ID or mugshot.  This is intended to enable recognition from more angles (especially elevation) as well as to support WB and gait recognition.  We took still photographs of subjects at two stations. Each station had three Nikon DSLR cameras mounted at different elevation angles that were configured to trigger simultaneously for each pose. The first station (shown in Figure~\ref{fig:BGC1_physical_layout}-1) captured passport-quality (i.e., neutral) face images from elevation angles of 0\textdegree, 20\textdegree, and 30\textdegree and yaw angles of -90\textdegree, -45\textdegree, 0\textdegree, 45\textdegree, and 90\textdegree with respect to the camera. The second station (Figure~\ref{fig:BGC1_physical_layout}-2 captured WB images from elevation angles of 0\textdegree, 20\textdegree, and 50\textdegree, and a full 360\textdegree rotation of yaw angles in increments of 45\textdegree. Examples of these images can be seen in Figure \ref{fig:controlled_example}. After completing both image stations, each subject then proceeded to a semicircular area with 10 indoor close-range bullet-style security cameras arranged around its arc to collect gait data: 
\begin{itemize}
    \item \textbf{Structured walking} -- The subject completed two back-and-forth passes along a straight line spanning the diameter of the semicircle.
    \item \textbf{Random walking} -- The subject walked a random path within the semicircle for 20 seconds, pretending to text and make a phone call during the second half of the 60 second period.
\end{itemize}
In the first collection event, the controlled portion took place in a large tent to simulate an indoor setting. In subsequent events, this portion took place indoors in a lab space and a conference room.

\subsection{Field Collection}

All field collection activities took place inside a 10 m $\times$ 10 m square area with cones marking the corners and midpoints of each boundary, as seen in \ref{fig:close_range_example} and~\ref{fig:BGC1_physical_layout} 5\&6. Colored lines intersected in the middle of the area and extended to the cones. Subjects performed the following three activities: 
\begin{itemize}
    \item \textbf{Standing} -- The subject stood in the center and rotated incrementally such that they faced along each marked line for 10 seconds.
    \item \textbf{Structured walking} -- The subject walked back and forth along each colored line one time to get videos from eight different directions.
    \item \textbf{Random walking} -- The subject walked a random path within the square for 60 seconds, pretending to send text messages and make a phone call during the latter half of the 60 second period. 
\end{itemize}
 
\subsubsection{Range Stations}

\begin{figure}
    \centering
    \includegraphics[scale=0.2]{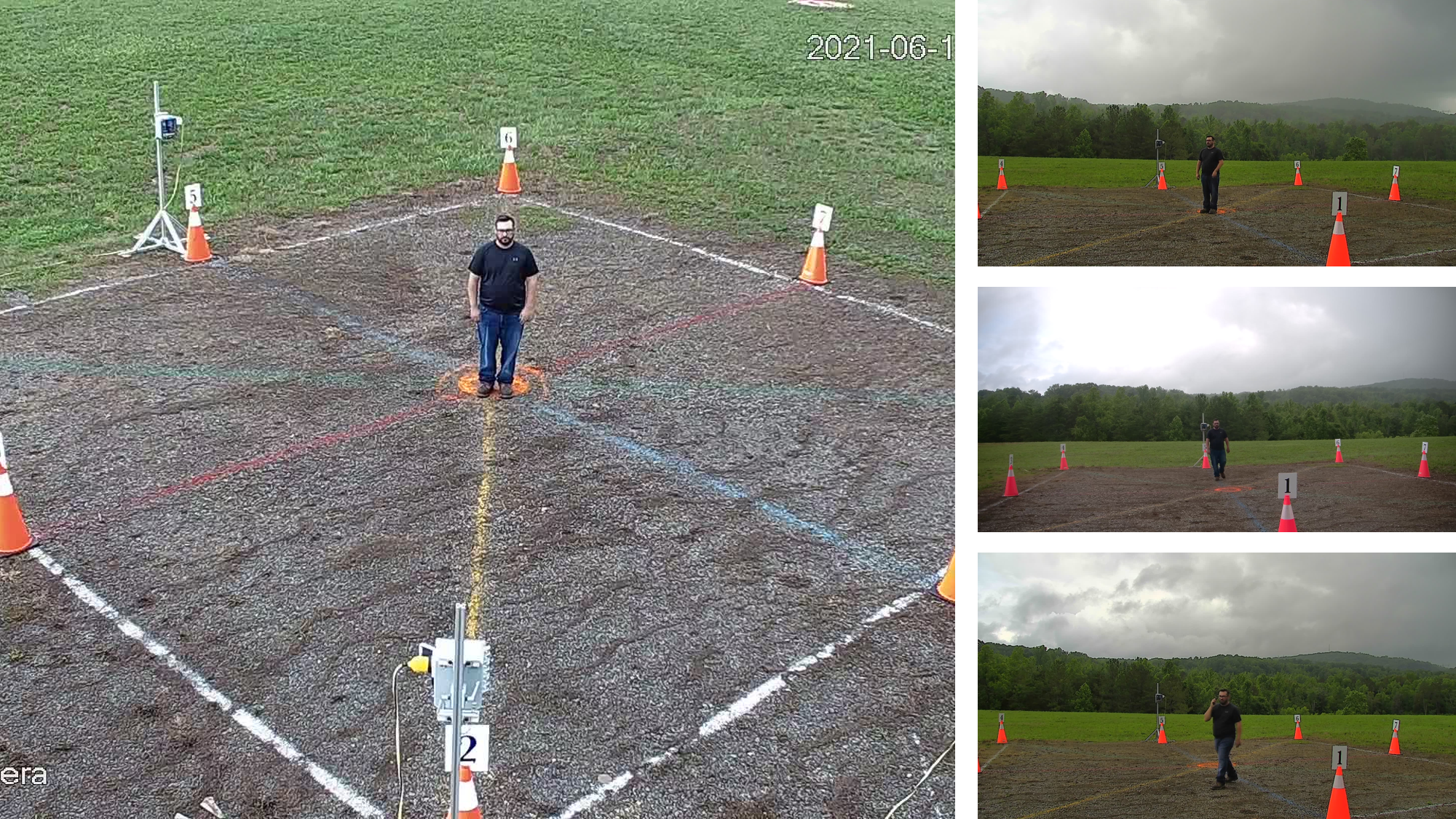}
    \caption{Close-range station example from BGC1. The image on the left is a single frame collected from an elevated surveillance camera. The images on the right are frames from the standing and walking portions of the subject collection viewed from ground-level cameras.}
    \label{fig:close_range_example}
\end{figure}

Cameras were grouped into stations designated by distance in meters from the field collection box. They were mounted on tripods on the ground, elevated on masts, or installed on the rooftops of buildings. Unlike the controlled collection outlined in section~\ref{subsec:controlled_collection}, the field collection layout differed between collection events due to the terrain and layout of the sites. Stations were laid out in a best-fit design to ensure the best use of range within the hosting location and economy of power sources. Ranges of up to 1000m are represented in the dataset, though the specific ranges selected for each collection differed. In addition, multiple cameras were deployed along the front edge of the collection box. These included surveillance cameras for high-resolution close-range imaging and Intel RealSense depth cameras for tracking subjects' movements and positions across time. Several surveillance cameras at close-range were mounted to a mast to capture the subject box from 30\textdegree and 50\textdegree pitch angles. Details of the close-range sensor layout can be seen in Figure \ref{fig:close_range_example}.

Range cameras were categorized as follows:
\begin{itemize}
    \item \textbf{Commercial surveillance cameras} -- We selected cameras that cost between \$500 and \$10k and represent a range of quality typical of standard security systems. Some of the highest-quality cameras were placed up to 500m from the subject box, which may exceed their usable range for biometric identification.
    \item \textbf{Military-grade cameras} -- These are typical of high-end and long-range military or border control cameras in a price range from \$100k to \$1.5M. These are built for very-long-range surveillance but not necessarily intended for biometric recognition.  These were deployed at distances of 500m to 1,000m.
    \item \textbf{Specialized cameras} -- These cameras were custom-built for this research. While the purchase price is typically less than military-grade cameras, they do contain sensors and optics with similar characteristics. These camera systems were designed with BRIAR program goals in mind and are used to explore key research topics. Sensors use global shutters, both color and grayscale configurations, and shorter integration times to help with turbulence and motion blur. Optics are typically high-end and carefully selected to provide high-quality images from 200m to 1,000m distances. These cameras also record two video streams: a high-quality compressed H264 stream and a lossless video stream to better support image enhancement research. 
\end{itemize}

The BGC1 collection was held at the same location at two different times of year. Additional systems were deployed for the second part of the collection. The first portion of the BGC1 collection comprised four range stations at 100m, 200m, 400m, and 500m distances from the center of the subject box. Two stations were deployed on either side of the subject box, allowing for two different view angles. The second portion of the BGC1 collection added a 300m imaging station with the same view angle as the 400m and 500m stations. The top-down view of the BGC1 deployment can be seen in Figure \ref{fig:BGC1_physical_layout}.

The BGC2 collection event included more stations and cameras, with the furthest distance from the subject measuring 1,000m. The layout of the collection site also required that the range stations be oriented to the subject field box from multiple view angles, some of which were placed on rooftops. Seven range stations deployed in BGC2. The 100m and 800m stations consisted entirely of cameras mounted to ground-based tripods.  The 600m station consisted of both cameras mounted to ground-based tripods as well as mast-mounted cameras.  The 1,000m stations consisted of a single mast-mounted system.  The 270m, 370m, and 490m stations were deployed on rooftops. The high variability in view angles and deployment methods ultimately led to a more diverse dataset in terms of image quality and impacts by atmospheric perturbations. The top-down view of the BGC2 deployment can be seen in Figure \ref{fig:BGC2_physical_layout}.

\subsubsection{UAV Collection}

We collected data using a variety of UAV platforms ranging in size from small quadcoptors to moderately sized Group 2 fixed-wing aircraft. The platforms were selected to represent operationally relevant aircraft similar to those used by the military or law enforcement agencies. In some cases, civilian proxies were used when there were difficulties obtaining some platforms. A list of aircraft used in the collections is shown in Table \ref{tab:aircraft}. On a typical collection day, the R80D, Puma, and SuperVolo were flown to cover larger aircraft while one or two of the smaller quadcopters flew patterns at low altitude near the collection box.

The platforms represent a variety of sizes and designs to simulate a number of scenarios. Up to four aircraft operated simultaneously during the collection and were typically separated at a range of altitudes and distances from the collection box. Flight time was an important consideration for the platforms; whereas larger aircraft could operate for hours at a time, the smaller quadcopters needed to be swapped every few subjects to change batteries. Keeping all aircraft operating proved to be very challenging due to frequent issues with battery swaps, maintenance, and weather. 

\begin{table*}
\centering
\begin{tabular}{llllll}
\toprule
Manufacturer & Name & Payload & Type & Typical altitude & Flight time \\
\midrule
Skydio & X2 & -- & Quadcopter & 50 ft & 0.5 hr \\
Autel  & Evo II Dual & -- & Quadcopter & 50 ft & 0.6 hr \\
Parrot & Anafi & -- & Quadcopter & 50 ft & 0.75 hr \\
Parrot & Anafi USA & -- & Quadcopter & 50 ft & 0.75 hr \\
FLIR & R80D & Zoom 30 or EO/IR MK-II & Quadcopter & 200 ft & Tethered \\
Aerovironment & Puma & MANTIS i45 & Fixed Wing & 600 ft & 2.5 hr \\
The Hybrid Project & SuperVolo & Trillium HD45 & Fixed Wing & 1,200 ft & 8.0 hr\\
\bottomrule
\end{tabular}

\caption{A list of aircraft used in BRIAR collections}
\label{tab:aircraft}
\end{table*}

\section{Dataset Curation and Organization}
\label{sec:dataset}

Once collected, data is separated into the BRIAR Research Set and the BRIAR Test Set. A stratified partitioning was performed across race, age, gender, body mass index, and height. Both partitions were provided to the Performer teams with the understanding that no subjects from the test set were to be used in training datasets. Raw recordings were cut into segments such that each final video included one subject wearing one clothing set performing one activity. 

\subsection{Metadata}

Metadata for each piece of BRIAR media is documented in an accompanying file in a standardized biometrics XML format \cite{ISO30137-4}. Though the availability and relevance of certain metadata varies across individual sensors and sensor types, all media XML files contain details about the piece of media itself such as resolution, frame rate and duration, demographic and media-specific information about the subject, and sensor details. Metadata for field videos includes environmental information such as venue descriptions and metrics describing atmospheric conditions. 

\subsection{Annotation}

Metadata for videos also includes annotations, which were generated using a combination of automated methods and manual efforts. Using RealSense cameras and elevated views, it is possible to track the subject’s position within the field collection box. This provides an estimate for the video sequence and identifies the different segments of the standing and structured walking activities. Directions during these segments are known and can be used to label with accuracy the expected yaw angle relative to each camera position. The assumption is that the subject’s pose stays the same at the estimated yaw angle throughout that segment. This activity timing metadata was also made available in some cases.  

\subsubsection{Automatic Annotation}
A high-level description of the initial automated annotation method is presented here. Annotations consist of non-rotated bounding boxes localizing the face and whole body of subjects in each video as well as yaw and pitch angles of the detected face. The face bounding box attempts to follow the annotations within the IJB-C (IARPA Janus Benchmark C), where the bounding box encapsulates the entire head if a significant part of the face is visible \cite{IJBC}. The WB bounding boxes intuitively capture the entire extent of a person such as their limbs even when extended. Yaw and pitch angles of the head denote a side–side and up–down movement, respectively. An exactly frontal image of a head would have a 0\textdegree yaw and pitch. 
\label{subsubsubsec:detection_method}
The current methodology for detections involved using YOLOv5 and the InsightFace package \cite{yolov5, scrfd}. Both are popular high-performing detection models trained on large datasets, so they were deemed suitable for the wide variety of environments to be encountered in the BRIAR data collection events. Face and WB detection was performed at 0.1 second intervals to achieve higher throughput due to the dataset size. Initially, annotation was performed utilizing software frameworks such as FaRO~\cite{bolme2020face,brogan2022faro} and the BRIAR specific API~\cite{brogan2022ornl} to quickly deploy automatic annotation pipelines. However, as annotations became more fine-grained and specific, use of these general frameworks have since been replaced by narrower, more powerful annotation workflows, detailed below.

Ideally, the subject should be the only person visible; however, other study team members were observed in multiple videos. It is very challenging to keep extra people out of view, especially in the UAV videos, given the constant motion of the vehicle and the wide field of view.  Therefore, DG-Net++ was utilized to compute a similarity score for each WB detection against the most appropriate controlled image available \cite{zou2020joint}. In the final datasets, these extra people were removed by placing boxes over their image.

Yaw angles were computed based on 2D joints extracted from either the constituent video or from a close-range video that was recorded simultaneously \cite{Zhang_2020_CVPR}. These yaw angles were used to find the most similar controlled image for comparison with DG-Net++. Tracklets were generated for each initial bounding box using BYTE and filtered by the similarity score statistics \cite{zhang2022bytetrack}. 

Most face detection models produce close crops of the face, which does not follow the style used in the IJB-C dataset. Therefore, 3D facial landmarks were generated with the "3d68" model from the InsightFace project. A set of heuristics based on the landmark points was used to output a bounding box encapsulating the entire head when a face is detected. Yaw and pitch angles were also derived from the same landmark points. Only overlapping face and WB bounding boxes were kept based on the WB detection processing mentioned previously in Section~\ref{subsubsubsec:detection_method}. 
 
\subsubsection{Manual Annotation}

Manual annotations for this dataset were used to verify the automatic annotations and fill critical gaps such as frames with challenging poses, partially occluded faces and bodies, and poor illumination conditions. Data was carefully selected for manual annotation to maximize impact and mitigate quality assurance issues due to resolution, extreme angles, ranges, and weather conditions. WB reference images captured in a controlled environment at a pitch angle of 0\textdegree ~were provided with videos of interest to verify subject and clothing set. In some cases, annotators were directed to perform additional screening for extraneous people in the background of videos to ensure that they were excluded from all released data. These cases were determined by finding WB tracklets that had temporal overlap in a video. The manual annotators are sent the minimum number of frame numbers to determine which of the WB detections correspond to the desired subject. Other detections are verified to be people or labeled as a misclassification.

\section{Dataset Characteristics}
\label{sec:characteristics}



We faced significant challenges to recruiting an ideal distribution of demographics across the subject pool. These challenges can be attributed to the regional demographics of eastern Tennessee, the collection location, and scheduling limitations. The collection site was not accessible by public transportation, and available appointments fell within standard business hours. BGC1 skews female, white, non-Hispanic, and older than age 50. We were able to improve the demographic imbalances somewhat in BGC2 by offering weekend participation and collecting in central Georgia, where the regional population is more diverse. Diverse representation across race, ethnicity, and gender is essential to combating performance differentials known to affect even the most accurate face recognition algorithms \cite{250171}. Given that a fundamental aim of the BRIAR program is to contribute a high-quality body of data that is inclusive of numerous covariates, eliminating disparities of this type is a top priority. We anticipate that by conducting subsequent collection events in different regions throughout the United States and recruiting participants with diversity in mind, we can build a demographically balanced dataset. 


Additionally, data collection endeavors as complex and ambitious as these are vulnerable to numerous points of failure. To date, some of these have included equipment failure, network inconsistency, inclement weather, and human error. Equipment failures encompass sensor malfunction, software instability, and mechanical defects or damage. Inclement weather can preclude UAV platforms from flying or cause them to crash, as well as introduce severe imaging restrictions. In particularly severe weather, it would be unsafe for researchers and subjects to proceed, although it slows down collection activities. Human error can adversely impact all aspects of data collection including proctor recording, equipment setup and usage, and inconsistent adherence to protocols over time. BGC1 and BGC2 contain, however sparsely, reflections of these impacts. The primary example of this is missing media files from certain sensors for certain subjects. Work is being done to mitigate in subsequent phases all adverse outcomes experienced during the first collection event, but none of these things is assuredly avoidable at all times.

\section{Conclusions and Future Work}
\label{sec:conclusion}

This dataset will be a groundbreaking addition to the field of biometrics and will enable advances in face and WB recognition previously hampered by the absence of robust, well-curated data targeting long-range and extreme pitch collections. 

Over the next three years, we will hold six more large collection events that will feature increasing complexity and naturalism. To the extent possible, each collection phase will take place in a different US region with unique environmental factors and a new set of subjects. Capturing data that represents varied terrain, climate and weather conditions, and considerable demographic representation is vital to the thorough development and rigorous evaluation of recognition algorithms intended for high-stakes use case scenarios. Ultimately, the goal for the completed dataset is participation by over 3,000 subjects. 

The final product will be a very large, rich dataset that can be used for a variety of experiments involving detection, enhancement, face and WB recognition, and reidentification.

\section{Acknowledgements and Copyright}

The authors would like to acknowledge the contributions and assistance of the following organizations and individuals in their support of the creation of the BRIAR dataset:

\paragraph{US Government Agencies:}

\begin{itemize}
    \item NIST: Mei Ngan and Patrick Grother
    \item Army C2ISR: Kevin Miller and Tim Williams
\end{itemize}

\paragraph{Additional Support:}

\begin{itemize}

    \item ORNL Support Staff: Margaret Smith, Amanda Mottern, Christy Gambrell, Michaela Martin, Chris Gibbs, and Willy Besancenez
    \item UCOR: Steve Nolan
    \item Guardian Centers of Georgia
    
\end{itemize}

\small{This research is based upon work supported by the Office of the Director of National Intelligence (ODNI), Intelligence Advanced Research Projects Activity (IARPA), via D20202007300010. The views and conclusions contained herein are those of the authors and should not be interpreted as necessarily representing the official policies, either expressed or implied, of ODNI, IARPA, or the U.S. Government. The U.S. Government is authorized to reproduce and distribute reprints for governmental purposes notwithstanding any copyright annotation therein.}

Notice:  This manuscript has been authored by UT-Battelle, LLC, under contract DE-AC05-00OR22725 with the US Department of Energy (DOE). The US government retains and the publisher, by accepting the article for publication, acknowledges that the US government retains a nonexclusive, paid-up, irrevocable, worldwide license to publish or reproduce the published form of this manuscript, or allow others to do so, for US government purposes. DOE will provide public access to these results of federally sponsored research in accordance with the DOE Public Access Plan (\url{http://energy.gov/downloads/doe-public-access-plan}).
{\small
\bibliographystyle{ieee_fullname}
\bibliography{egbib}

\begin{thebibliography}{10}\itemsep=-1pt

\bibitem{odnireport}
Principles of artificial intelligence ethics for the intelligence community,
  June 2020.

\bibitem{FAAlimit}
{FAA} part 107 fact sheet, 2022.

\bibitem{brogan2022faro}
Joel Brogan, Nell Barber, David Cornett, and David Bolme.
\newblock Faro 2: an open source, configurable smart city framework for
  real-time distributed vision and biometric systems.
\newblock {\em arXiv preprint arXiv:2209.12962}, 2022.

\bibitem{brogan2022ornl}
Joel Brogan, Bob Zhang, David Bolme, and Mei Ngan.
\newblock Ornl/briar-api.
\newblock Technical report, Oak Ridge National Lab.(ORNL), Oak Ridge, TN
  (United States), 2022.

\bibitem{Corona_2021_WACV}
Kellie Corona, Katie Osterdahl, Roderic Collins, and Anthony Hoogs.
\newblock Meva: A large-scale multiview, multimodal video dataset for activity
  detection.
\newblock In {\em Proceedings of the IEEE/CVF Winter Conference on Applications
  of Computer Vision (WACV)}, pages 1060--1068, January 2021.

\bibitem{bolme2020face}
Joel~Brogan David S.~Bolme, Nisha~Srinivas and David Cornett.
\newblock Face recognition oak ridge (faro): A framework for distributed and
  scalable biometrics applications.
\newblock In {\em 2020 IEEE International Joint Conference on Biometrics
  (IJCB)}, pages 1--8, 2020.

\bibitem{ISO30137-4}
International~Organization for Standardization.
\newblock {\em {Use of biometrics in video surveillance systems — Part 4:
  Ground truth and video annotation procedure}}.
\newblock International Organization for Standardization, Vernier, Geneva,
  Switzerland, {ISO}/{IEC} {TR} 30137-4:2021 edition, 2021.

\bibitem{250171}
Patrick Grother, Mei Ngan, and Kayee Hanaoka.
\newblock Face recognition vendor test part 3: Demographic effects, 2019-12-19
  2019.

\bibitem{scrfd}
Jia Guo, Jiankang Deng, Alexandros Lattas, and Stefanos Zafeiriou.
\newblock Sample and computation redistribution for efficient face detection.
\newblock {\em CoRR}, abs/2105.04714, 2021.

\bibitem{gunther_unconstrained_2017}
Manuel Günther, Peiyun Hu, Christian Herrmann, Chi~Ho Chan, Min Jiang, Shufan
  Yang, Akshay~Raj Dhamija, Deva Ramanan, Jürgen Beyerer, Josef Kittler,
  Mohamad~Al Jazaery, Mohammad~Iqbal Nouyed, Guodong Guo, Cezary Stankiewicz,
  and Terrance~E. Boult.
\newblock Unconstrained {Face} {Detection} and {Open}-{Set} {Face}
  {Recognition} {Challenge}.
\newblock In {\em 2017 {IEEE} {International} {Joint} {Conference} on
  {Biometrics} ({IJCB})}, pages 697--706, Oct. 2017.
\newblock arXiv:1708.02337 [cs].

\bibitem{yolov5}
Glenn Jocher, {Ayush Chaurasia}, Alex Stoken, Jirka Borovec, {NanoCode012},
  {Yonghye Kwon}, {TaoXie}, {Kalen Michael}, {Jiacong Fang}, {Imyhxy}, {,
  Lorna}, Colin Wong, {(Zeng Yifu)}, {Abhiram V}, Diego Montes, {Zhiqiang
  Wang}, Cristi Fati, {Jebastin Nadar}, {Laughing}, {UnglvKitDe}, {Tkianai},
  {YxNONG}, Piotr Skalski, Adam Hogan, Max Strobel, Mrinal Jain, Lorenzo
  Mammana, and {Xylieong}.
\newblock ultralytics/yolov5: v6.2 - yolov5 classification models, apple m1,
  reproducibility, clearml and deci.ai integrations, 2022.

\bibitem{Kalka_MDF}
Nathan~D. Kalka, James~A. Duncan, Jeremy Dawson, and Charles Otto.
\newblock Iarpa janus benchmark multi-domain face.
\newblock In {\em 2019 IEEE 10th International Conference on Biometrics Theory,
  Applications and Systems (BTAS)}, pages 1--9, 2019.

\bibitem{Kalka_IJBS}
Nathan~D. Kalka, Brianna Maze, James~A. Duncan, Kevin O’Connor, Stephen
  Elliott, Kaleb Hebert, Julia Bryan, and Anil~K. Jain.
\newblock Ijb–s: Iarpa janus surveillance video benchmark.
\newblock In {\em 2018 IEEE 9th International Conference on Biometrics Theory,
  Applications and Systems (BTAS)}, pages 1--9, 2018.

\bibitem{januspaper}
Brendan~F. Klare, Ben Klein, Emma Taborsky, Austin Blanton, Jordan Cheney,
  Kristen Allen, Patrick Grother, Alan Mah, and Anil~K. Jain.
\newblock Pushing the frontiers of unconstrained face detection and
  recognition: Iarpa janus benchmark a.
\newblock In {\em Proceedings of the IEEE conference on computer vision and
  pattern recognition}, pages 1931--1939, 2015.

\bibitem{pdestre}
S.~V.~Aruna Kumar, Ehsan Yaghoubi, Abhijit Das, B.~S. Harish, and Hugo
  Proença.
\newblock The p-destre: A fully annotated dataset for pedestrian detection,
  tracking, re-identification and search from aerial devices, 2020.

\bibitem{LFWTechUpdate}
Gary B. Huang~Erik Learned-Miller.
\newblock Labeled faces in the wild: Updates and new reporting procedures.
\newblock Technical Report UM-CS-2014-003, University of Massachusetts,
  Amherst, May 2014.

\bibitem{li2016toward}
Pei Li, Joel Brogan, and Patrick~J Flynn.
\newblock Toward facial re-identification: Experiments with data from an
  operational surveillance camera plant.
\newblock In {\em 2016 IEEE 8th International Conference on Biometrics Theory,
  Applications and Systems (BTAS)}, pages 1--8. IEEE, 2016.

\bibitem{liu2015faceattributes}
Ziwei Liu, Ping Luo, Xiaogang Wang, and Xiaoou Tang.
\newblock Deep learning face attributes in the wild.
\newblock In {\em Proceedings of International Conference on Computer Vision
  (ICCV)}, December 2015.

\bibitem{9432821}
Luis~S. Luevano, Leonardo Chang, Heydi Méndez-Vázquez, Yoanna
  Martínez-Díaz, and Miguel González-Mendoza.
\newblock A study on the performance of unconstrained very low resolution face
  recognition: Analyzing current trends and new research directions.
\newblock {\em IEEE Access}, 9:75470--75493, 2021.

\bibitem{IJBC}
Brianna Maze, Jocelyn Adams, James~A. Duncan, Nathan Kalka, Tim Miller, Charles
  Otto, Anil~K. Jain, W.~Tyler Niggel, Janet Anderson, Jordan Cheney, and
  Patrick Grother.
\newblock Iarpa janus benchmark - c: Face dataset and protocol.
\newblock In {\em 2018 International Conference on Biometrics (ICB)}, pages
  158--165, 2018.

\bibitem{ristani_performance_2016}
Ergys Ristani, Francesco Solera, Roger~S. Zou, Rita Cucchiara, and Carlo
  Tomasi.
\newblock Performance {Measures} and a {Data} {Set} for {Multi}-{Target},
  {Multi}-{Camera} {Tracking}.
\newblock Technical Report arXiv:1609.01775, arXiv, Sept. 2016.
\newblock arXiv:1609.01775 [cs] type: article.

\bibitem{vidal2018ug}
Rosaura~G Vidal, Sreya Banerjee, Klemen Grm, Vitomir Struc, and Walter~J
  Scheirer.
\newblock Ug\^{} 2: A video benchmark for assessing the impact of image
  restoration and enhancement on automatic visual recognition.
\newblock In {\em 2018 IEEE Winter Conference on Applications of Computer
  Vision (WACV)}, pages 1597--1606. IEEE, 2018.

\bibitem{Zhang_2020_CVPR}
Feng Zhang, Xiatian Zhu, Hanbin Dai, Mao Ye, and Ce Zhu.
\newblock Distribution-aware coordinate representation for human pose
  estimation.
\newblock In {\em IEEE/CVF Conference on Computer Vision and Pattern
  Recognition (CVPR)}, June 2020.

\bibitem{zhang_person_2021}
Shizhou Zhang, Qi Zhang, Yifei Yang, Xing Wei, Peng Wang, Bingliang Jiao, and
  Yanning Zhang.
\newblock Person {Re}-identification in {Aerial} {Imagery}.
\newblock {\em IEEE Transactions on Multimedia}, 23:281--291, 2021.
\newblock arXiv:1908.05024 [cs].

\bibitem{zhang2022bytetrack}
Yifu Zhang, Peize Sun, Yi Jiang, Dongdong Yu, Fucheng Weng, Zehuan Yuan, Ping
  Luo, Wenyu Liu, and Xinggang Wang.
\newblock Bytetrack: Multi-object tracking by associating every detection box.
\newblock In {\em Proceedings of the European Conference on Computer Vision
  (ECCV)}, 2022.

\bibitem{zheng_rpifield_2018}
Meng Zheng, Srikrishna Karanam, and Richard~J. Radke.
\newblock {RPIfield}: {A} {New} {Dataset} for {Temporally} {Evaluating}
  {Person} {Re}-identification.
\newblock In {\em 2018 {IEEE}/{CVF} {Conference} on {Computer} {Vision} and
  {Pattern} {Recognition} {Workshops} ({CVPRW})}, pages 1974--19742, Salt Lake
  City, UT, USA, June 2018. IEEE.

\bibitem{zou2020joint}
Yang Zou, Xiaodong Yang, Zhiding Yu, BVK~Vijaya Kumar, and Jan Kautz.
\newblock Joint disentangling and adaptation for cross-domain person
  re-identification.
\newblock In {\em Computer Vision--ECCV 2020: 16th European Conference,
  Glasgow, UK, August 23--28, 2020, Proceedings, Part II 16}, pages 87--104.
  Springer, 2020.

\end{thebibliography}
}

\end{document}